%% file: main.tex
\documentclass[pmlr]{jmlr}

\input{preamble}

\begin{document}

\setlength{\abovedisplayskip}{6pt}  % space above
\setlength{\belowdisplayskip}{6pt}  % space below

\maketitle

\vspace{-20pt}

\begin{abstract}
Many tasks require mapping continuous input data (e.g. images) to discrete task outputs (e.g. class labels). Yet, how neural networks learn to perform such \textit{discrete computations on continuous data manifolds} remains poorly understood. Here, we show that signatures of such computations emerge in the representational geometry of neural networks as they learn. By analysing the Riemannian pullback metric across layers of a neural network, we find that network computation can be decomposed into two functions: discretising continuous input features and performing logical operations on these discretised variables. Furthermore, we demonstrate how different learning regimes (rich vs. lazy) have contrasting metric and curvature structures, affecting the ability of the networks to generalise to unseen inputs. Overall, our work provides a geometric framework for understanding how neural networks learn to perform discrete computations on continuous manifolds.
\end{abstract}

\bigskip


\section{Introduction}\label{sec:intro}

Studying the geometry of manifolds of neural activation can help interpret how neural networks perform tasks \citep{chung2021neural}. Empirically, dimensionality reduction methods have been used to infer low-dimensional structure in the high-dimensional activity of biological \citep{pellegrino2024dimensionality} and artificial neural networks \citep{li2023emergent}. More recently, analytical methods stemming from Riemannian geometry have related these low-dimensional manifolds of neural activation to task computations \citep{hauser2017principles, pellegrino2025rnns}. These works share a common basis: the manifold hypothesis, which posits that the input data to a network lies on low-dimensional continuous manifolds \citep{fefferman2016testing}. 

A parallel line of work has tackled how neural networks learn low-dimensional features of their data inputs \citep{saxe2013exact}. Studying the gradient dynamics of neural networks has helped gain insights into their generalisation ability, sample complexity, and other facets of task learning \citep{damian2024computational, mousavineural, abbe2023sgd}. Importantly, in an effort to gain analytical tractability, and with some notable exceptions, this line of work has focused on relatively unstructured inputs (e.g. Gaussian or Boolean), with a teacher network or polynomial target outputs. This stands in contrast with the representational geometry typically studied in networks receiving inputs on low-dimensional manifolds, and which are trained to produce discrete class labels \citep{aubry2015understanding}. 

Here, we argue that using Riemannian geometric tools to formally study the representations that emerge as neural networks learn can provide insight into such discrete computations on continuous manifolds. Throughout, we study how multi-layer perceptrons learn to implement Boolean functions (e.g. XOR, AND) on a continuous manifold of inputs (e.g. a torus or a plane). We find that changes in the Riemannian metric over learning reveal the discretisation of the continuous input manifold in the early layers of the network and the discrete Boolean operation in subsequent layers.  Furthermore, rich and lazy learning regimes can lead to structured and random representational geometries, with different generalisation abilities. Finally, we show that input noise during training decreases the curvature of the manifold, and that it corresponds to the network learning a flatter posterior distribution of the target output. Overall, our work links continuous input data manifolds to discrete task output by studying the Riemannian geometry of hidden layer activation.

\subsection*{Contributions}

\begin{enumerate}[leftmargin=*, labelindent=0pt]

    \item \textbf{Riemannian geometry of discrete computation on continuous manifolds.} We introduce a framework anchored in Riemannian geometry to study discrete computations on continuous manifolds. We show in multi-layer perceptrons trained to implement Boolean functions that the metric tensor becomes highly localised close to class boundaries to reflect the discretisation of the continuous input manifold.

    \item \textbf{Different learning regimes have different representational geometries.} In the same model we show that feature learning partitions the computation into a binarisation corresponding to the collapse of the embedding space of the input manifold and a degenerate metric tensor, followed by the discrete logical computations. Furthermore, this geometry generalises to unseen inputs on the manifold, while lazy learning doesn't.

    \item \textbf{Noise smooths the geometry to implement Bayesian computation.} We show that the curvature of the manifold decreases with input noise level, which corresponds to a flatter posterior distribution of the output of the network.

\end{enumerate}

\subsection*{Related works}

\noindent \textbf{Feature learning.} Key work has derived the nonlinear learning dynamics of deep linear networks to show that they could learn in two qualitatively distinct ``rich'' (or ``feature'') and ``lazy'' (or ``kernel'') learning regimes \citep{saxe2013exact}. Since then, similar tools have been used to tackle various questions regarding how networks learn, such as understanding how many samples are required to learn target functions of increasing complexity \citep{dandi2023two} or bounding the generalisation error of neural networks \citep{goldt2019dynamics}. Importantly, these works often focus on Gaussian inputs or continuous target functions (polynomials or teacher networks). Here, we instead study feature learning in neural networks trained to map continuous input manifolds to discrete outputs.

\medskip

\noindent \textbf{Representational geometry of neural networks.} A long line of work has argued that geometrically studying how neural networks encode data features in their activations can help understand how they solve tasks, with applications in vision \citep{aubry2015understanding}, large language models \citep{li2023emergent}, and reinforcement learning \citep{tennenholtz2022uncertainty}. In particular, tools from Riemannian geometry have been used to characterise the exact intrinsic geometry of neural representation of networks receiving low-dimensional inputs \citep{hauser2017principles, Benefati2023}. Here, we investigate how learning discrete target outputs affects the Riemannian geometry of the hidden layer activation. 

\section{Riemannian geometry provides the tools to study the intrinsic geometry of neural network representations}

In this section we briefly review concepts from Riemannian geometry that we will use to study neural network representations. In particular, we will exactly characterise the intrinsic geometry of the hidden layer activation. We work under the manifold hypothesis, and the overall neural network can be summarised as:
\[
    \mathcal{M} \xrightarrow{\psi} \mathbb{R}^{n_\text{in}} \xrightarrow{\varphi} \mathbb{R}^n \xrightarrow{\zeta} \mathbb{R}^{n_\text{out}}
\]
The input data manifold is $\mathcal{M}$, which is embedded into the input space of the neural network $\mathbb{R}^{n_\text{in}}$ via $\psi$. Assuming that we are interested in studying the representational geometry of a particular layer, we can decompose the neural network into two functions: $\varphi$, which maps the embedded inputs to the hidden layer of interest, and $\zeta$ mapping this hidden layer to the output. Under weak constraints, the activation of the network in response to all possible inputs $(\varphi\circ \psi)(\mathcal{M})$ will itself be a manifold, of the same topology as $\mathcal{M}$. We are interested in characterising the geometry of this manifold as it sits in the hidden-layer state-space. 

To characterise the intrinsic geometry of the hidden layer activation, the pullback of the metric can be computed:
\[
    g:T_p\mathcal{M} \times T_p\mathcal{M} \xrightarrow{\mathrm{d}\psi,\mathrm{d}\psi}\mathbb{R}^{n_\text{in}}\times \mathbb{R}^{n_\text{in}}\xrightarrow{\mathrm{d}\varphi, \mathrm{d}\varphi} \mathbb{R}^n \times \mathbb{R}^n \xrightarrow{\langle \cdot,\cdot\rangle} \mathbb{R}
\] 
where $T_p\mathcal{M}$ is the tangent space at a point $p\in\mathcal{M}$. Intuitively, tangent vectors of the input manifold can be mapped to vectors tangent to the manifold in the input embedding space $\mathbb{R}^{n_\text{in}}$ via the pushforward of the input embedding $\mathrm{d}\psi$, and then to vectors tangent to the manifold in the hidden layer state-space $\mathbb{R}^n$ via the Jacobian of the neural network $\mathrm{d}\varphi$. Thus, the inner product between two tangent vectors of the input manifold can be measured via the standard Euclidean dot product $\langle \cdot, \cdot \rangle$ by pushing them forward to the relevant Euclidean state-space. 

This definition is independent of the choice of local coordinates on the manifold. However, in many cases, there will be a basis that represents the particular task variables at hand --- e.g. the orientation and translation of an object in image space --- whose neural representation we seek to understand. In this case, the metric can be summarised by how it acts on this basis: if we call $\mathbf{z}(p_1, p_2)\in\mathbb{R}^n$ the activation of the network in response to an input depending on two task variables $p_1, p_2\in\mathbb{R}$, we can characterise the encoding of these task variables by asking how the neural representation changes in response to small changes in the first $\partial_{p_1}\mathbf{z}=\frac{d\mathbf{z}(p_1, p_2)}{dp_1}$ or second $\partial_{p_2}\mathbf{z}=\frac{d\mathbf{z}(p_1, p_2)}{dp_2}$ variable. If variables have locally correlated representations, then changing them individually will cause similar changes in representation, whereas uncorrelated representations cause changes in orthogonal directions in neural state-space. Mathematically, this is equivalent to saying that $\partial_{p_1}\mathbf{z} \cdot \partial_{p_2}\mathbf{z} = 0$. In a task involving $k$ variables, the pullback metric summarises such correlations:
    \[
        G = \begin{bmatrix}
                \partial_{p_1}\mathbf{z} \cdot \partial_{p_1}\mathbf{z} & \hdots & \partial_{p_1}\mathbf{z}\cdot \partial_{p_k}\mathbf{z} \\
                \vdots &  \ddots & \vdots \\
                \partial_{p_k}\mathbf{z} \cdot \partial_{p_1}\mathbf{z}   & \hdots & \partial_{p_k}\mathbf{z} \cdot \partial_{p_k}\mathbf{z}              
            \end{bmatrix}
    \]
Studying this pullback metric, and therefore the local correlations in the encoding of different task variables, can provide insights into the computations performed by a network.

\section{Neural network models can perform discrete computations on continuous manifolds of inputs}

Classic work has shown that neural networks can approximate arbitrary Boolean functions \citep{siegelmann1992computational}. In this work we consider a continuous extension of this task, involving learning Boolean functions, while allowing inputs to lie on a low-dimensional manifold. This provides a minimal yet rich setting for analysing how representational geometry supports discrete computations on continuous manifolds. More specifically, we start with a simple example where the input manifold is a so-called \textit{flat torus}, consisting of the Cartesian product of two circles, each encoding a different input to the Boolean function. To generate the input manifold, we define two angular task variables, $\theta_1,\theta_2\in[0,2\pi)$, which form a torus embedded in $\mathbb{R}^4$:
\[
    \mathbf{x}=\begin{bmatrix}\cos(\theta_1) & \sin(\theta_1) & \cos(\theta_2) & \sin(\theta_2)\end{bmatrix}^\top.
\]
We define a class boundary on each input circle at an angle $\alpha$, such that $0\leq\theta-\alpha<\pi$ is mapped to $+1$, and $\pi\leq\theta-\alpha<2\pi$ is mapped to $0$. The network is trained to apply a Boolean function (e.g. XOR) on the discretised variables.

\subsection{The learning dynamics of linear networks reveal that task-specific representations emerge in hidden layer activation}\label{sec:learning_dynamics}

We first consider the learning dynamics of a linear network performing the AND task. Without loss of generality we can set $\alpha=0$ and recover solutions for all other boundary rotations with a rotation of the input (Appendix~\ref{apd:weight_rotation}). The target output is:
\[
y(\theta) =\quad
1 \text{ if } (0 \leq \theta_1 < \pi \text{ and } 0 \leq \theta_2 < \pi)\quad 0\text{ else}
\]
We consider a network which is linear in $\mathbf{x}$, but not $\theta$:
\[
y=W_2W_1\mathbf{x}(\theta).
\]
By extending a framework developed in \cite{saxe2013exact} to consider inputs on a manifold, we can derive exact learning dynamics for the linear network in terms of correlation matrices. The input-input and input-output correlations are defined by the input embedding:
\[
\Sigma^{11}=\mathbb{E}[\mathbf{x}\mathbf{x}^\top]= 0.5\mathbf{I}, \quad\quad \Sigma^{31} = \mathbb{E}[y\mathbf{x}^\top] = \begin{bmatrix}0&\frac{1}{2\pi}&0&\frac{1}{2\pi}\end{bmatrix}
 \]
Under certain assumptions (detailed in Appendix \ref{apd:saxe_dynamics}), the dynamics across training of the learned modes are controlled by the singular value decomposition of $\Sigma^{31}=USV^\top$. For the AND task with input embedding as above, there is exactly one non-zero singular value $s_1=1/\pi\sqrt{2}$ with corresponding right singular vector $\mathbf{v}_1=\begin{bmatrix}0 & 1/\sqrt{2} & 0 & 1/\sqrt{2}\end{bmatrix}^\top$.

It follows that the network will learn exactly one mode corresponding to $m_1 = \mathbf{v}^\top_1\mathbf{x}\propto\sin{\theta_1}+\sin{\theta_2}$. The network weights corresponding to the task relevant inputs (sine of the input angles) evolve jointly, whilst the weights mapping the task irrelevant inputs (cosine of the input angle) are not updated, resulting in a hidden weight structure of the form:
\[
W_1=[\mathbf{0}\,\, \mathbf{a} \,\,\mathbf{0}\,\, \mathbf{a}]
\]
where $\mathbf{a}=W_1\mathbf{v}_1$ is the projection of the hidden weights onto the learned mode.

The projection of the total weights onto the learned mode, $u=W_2W_1\mathbf{v}_1$, will evolve according to:
\vspace{-6pt}
\[
u(t) = \frac{S e^{St/\tau}}{e^{St/\tau} - 1 + \frac{S}{u_0}}
\]
where $S=2s_1$, $t$ is the training iteration and $\tau=\frac{1}{\text{Learning Rate}}$ is the time constant of learning. Figure~\ref{fig:saxe_learning_dynamics} shows that $u(t)$ accurately predicts learning dynamics in a linear network trained on the AND task. Thus, networks learn hidden layer representations that capture task-relevant information on the manifold via low-rank weights. 

Using the evolution of the weights, we can derive a similar expression for the dynamics of the hidden pullback metric:
\vspace{-6pt}
\[
G_\mathbf{z}(t)\simeq\frac{1}{2}a(t)^2 G_\mathbf{z}^\text{task} + G_\mathbf{z}^\perp
\] where $a(t)$ represents the scalar learning dynamics of $W_1$, $G_\mathbf{z}^\text{task}$ has structure corresponding to the task features and $G_\mathbf{z}^\perp$ is the metric derived from weights orthogonal to the learned mode at initialisation. Initially, $a(t)^2$ is small, and the metric is determined by random initialisations. In particular, in the limit of a large hidden layer, we obtain that at initialisation $G_\mathbf{z}(0)=I_2$, that is the manifold is uniform ($G_\mathbf{z}$ does not depend on $\theta_1,\theta_2$). Over learning $a(t)^2$ grows, dominating the $G_\mathbf{z}^\perp$ term so the metric becomes task-specific. More specifically, after training $G_\mathbf{z}(t) \simeq 0.5a(t)^2[[\cos^2 \theta_1, \cos \theta_1\cos \theta_2], [\cos \theta_1\cos \theta_2, \cos^2 \theta_2]]$, that is the manifold is non-uniform and the metric close to degenerate at some points.

Thus, we have shown that task-specific Riemannian geometry emerges over learning discrete computations on continuous input manifolds. Yet, here the network is linear and cannot exactly discretise the input variables or perform more complex Boolean operations, due to not being able to induce curvature on the manifold. Hence, in the next section we turn to studying nonlinear networks, trained on more complex tasks.

% ===== XOR model =====

\subsection{The metric and curvature of the representations of nonlinear networks tie continuous inputs to discrete task outputs}

\begin{figure}[h]
    \centering
    \includegraphics[width=\linewidth]{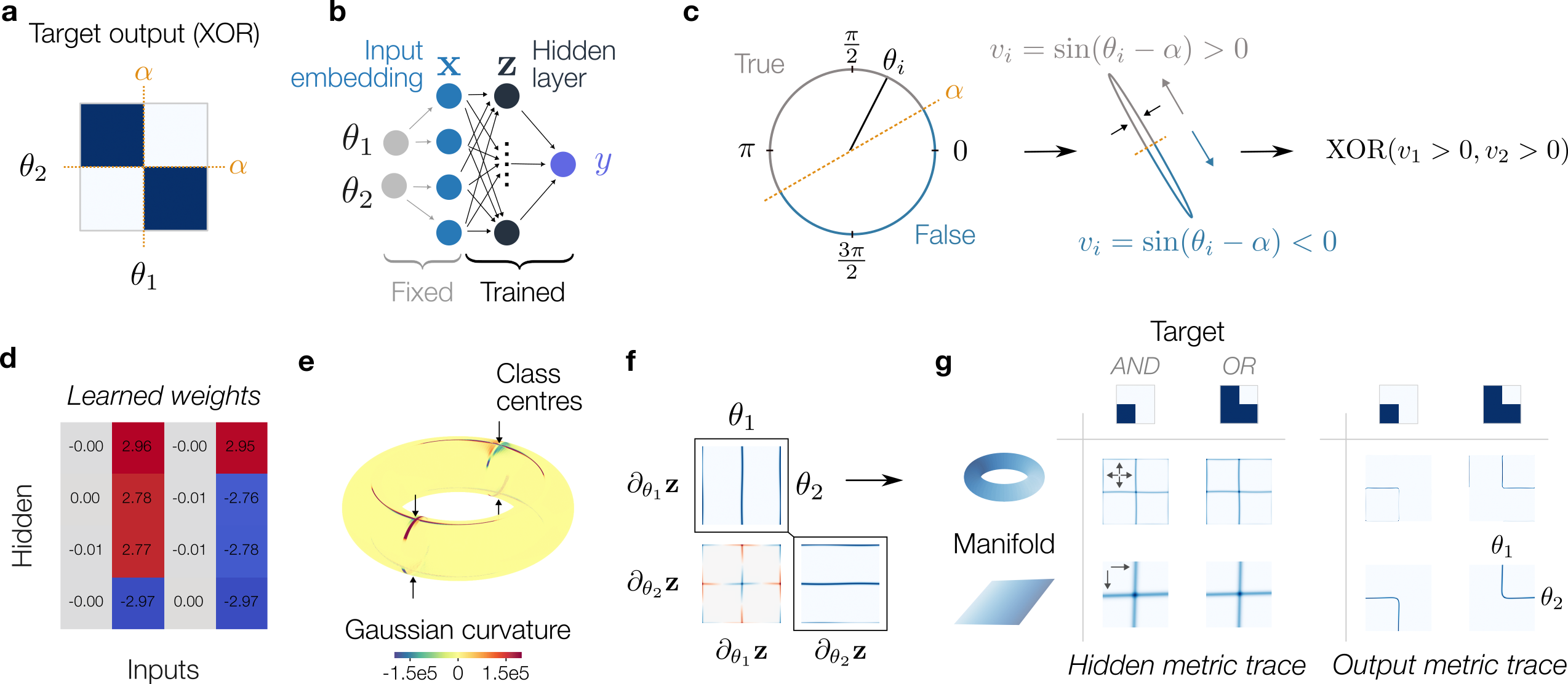}
    \caption{\textbf{Neural network geometry reflects discrete computations on manifolds.} \textbf{a.} The target outputs of the XOR task with two angular inputs $\theta_1,\theta_2\in[0,2\pi)$. 
    \textbf{b.} Diagram of the network architecture used to solve the task. 
    \textbf{c.} 
    \textit{Left:} Schematic showing an input variable $\theta_i$ embedded on a unit circle, $\mathbf{x}=[\cos(\theta_i),\sin(\theta_i)]^\top$, with decision boundaries at $\alpha$ and $\pi+\alpha$. 
    \textit{Right:} The optimal solution compresses the irrelevant dimensions and performs the logic operation on the 1D representation. \textbf{d.} Input weight matrix of a trained network (for $\alpha=0)$. Weights corresponding to $\mathbf{x}_1=\cos(\theta_1)$ and $\mathbf{x}_3=\cos(\theta_2)$ are close to zero.
    \textbf{e.} The Gaussian curvature of the hidden layer manifold visualised on a torus. The curvature diverges if $\theta_i\in\{\frac{\pi}{2},\frac{3\pi}{2}\}$. 
    \textbf{f.} The components of the metric tensors of the hidden layer activation of the network (lower-triangular part of the metric, each entry varies over the manifold). 
    \textbf{g.} Trace of the metric for networks trained on different combination of input manifolds (torus, plane) and target output (AND, OR).}
    \label{fig:fig1}
    \vspace{-3ex}
\end{figure}

We next consider the toroidal inputs generalisation of the XOR task. The network is trained to perform XOR on the discretised inputs, creating four quadrants (with periodic boundaries) on the input manifold (Fig.~\ref{fig:fig1}a). Although inputs to the network are 4-dimensional, an efficient representation is sensitive only to the sign of the following latent variables:
\[
    v_i(\theta_i,\alpha)=\sin(\theta_i-\alpha)
\]
such that the target output is $y=\text{XOR}[v_1(\theta_1,\alpha)>0,v_2(\theta_2,\alpha)>0]$ (Fig.~\ref{fig:fig1}c). We train a multi-layer perceptron with four hidden neurons (Fig.~\ref{fig:fig1}b) and tanh activation, using gradient descent on the binary cross-entropy loss.

We find that in networks trained on the XOR task on the flat torus, the input weights compress task-irrelevant dimensions. For example, when $\alpha=0$, the task output is independent of $x_1$ and $x_3$: the cosines of the two input angles (Fig.~\ref{fig:fig1}c-d). For $\alpha\neq 0$, the different linear combinations of the $x_i$'s will encode the task-relevant information. 
Thus, two dimensions of the input torus are collapsed, with the remaining inputs (e.g. $x_2,x_4\in[-1,1]$ for $\alpha=0$) defining a patch of the $\mathbb{R}^2$ plane. This is reflected by the intrinsic curvature of the torus in the hidden layer, which for collapsed $x_1$ and $x_3$ directions is approximately:
\[
    K\simeq\frac{M(\theta_1,\theta_2)}{ \cos^4(\theta_1)\cos^4(\theta_2)}
\] where $M(\theta_1,\theta_2)$ is a term depending on the first and second derivatives of the metric tensor.
When $\cos(\theta_i)=0$ for either $i$ the curvature diverges. This happens near the class centres $\theta_i\in\{\frac{\pi}{2},\frac{3\pi}{2}\}$ (Fig.~\ref{fig:fig1}e), where the circles are ``folded'' (Fig.~\ref{fig:fig1}c).

From the standpoint of Riemannian geometry, the hidden layer metric tensor encodes this effect. Visualising the entries of the metric tensor in the input coordinates highlighted that space near decision boundaries is stretched, while space far from them is compressed (Fig.~\ref{fig:fig1}f). This effectively discretises the inputs, with transitions between the discretised variables only occurring across boundaries. In comparison, the metric pulled back from the output layer reflects the logical gate  (Fig.~\ref{fig:fig1}g).
To see how this result depended on the choice of task and input manifold, we repeated this analysis in the AND and OR tasks, with inputs either on a torus or a plane. Specifically, we looked at the trace of the metric tensor, which reflects the overall stretching and compression of space at any point on the manifold. We found that the hidden layer metric's trace showed stretching of space near the input class boundaries, while the output metric showed stretching of space near the target output decision boundaries (Fig.~\ref{fig:fig1}g). Thus, the metric and curvature of the hidden layer reflect computations that are tied to the choice of input manifold and its embedding, while the output layer geometry reflects Boolean function-specific computation. 

Finally, we explored what happens when continuous and discrete computations cannot be partitioned across layers. To do that, we went back to the XOR task and compared the geometry of a 1-layer network to that of a 2-layer network. Classic work has shown that, with binary inputs to the network, the XOR task requires at least one hidden layer \citep{minsky1969introduction}. Thus, our 1-layer network cannot use its hidden layer to discretise the input manifold, while the 2-layer network can in principle discretise the input manifold in the first hidden layer before performing the classic solution on the binarised variables in the second layer. Despite this discrepancy, both 1- and 2-layer networks can perform the XOR task to a similar performance level  (Fig.~\ref{sup_fig1}a). Interestingly, the two networks used different geometries: the 2-layer network employed a similar representation of the discretised variable in the first layer, while the 1-layer network showed a rich geometry which combined the discretisation and Boolean computation  (Fig.~\ref{sup_fig1}b).

Overall, this suggests that computations in these simple networks are partitioned across layers into the discretisation of inputs manifolds and discrete Boolean computations.

\subsection{Rich and lazy learning of input manifold features result in different intrinsic geometries}

Next, we tried to better understand how different learning regimes affect the hidden layer representation and the overall discrete computations on the continuous manifold. Previous work has shown that the variance of the weights at initialisation can qualitatively affect the learned representation \citep{saxe2013exact}. Here we replicate these analyses (Fig. \ref{sup_fig2}) with our model of discrete computation on continuous non-linear manifolds.

\begin{figure}[h]
    \centering
    \includegraphics[width=\linewidth]{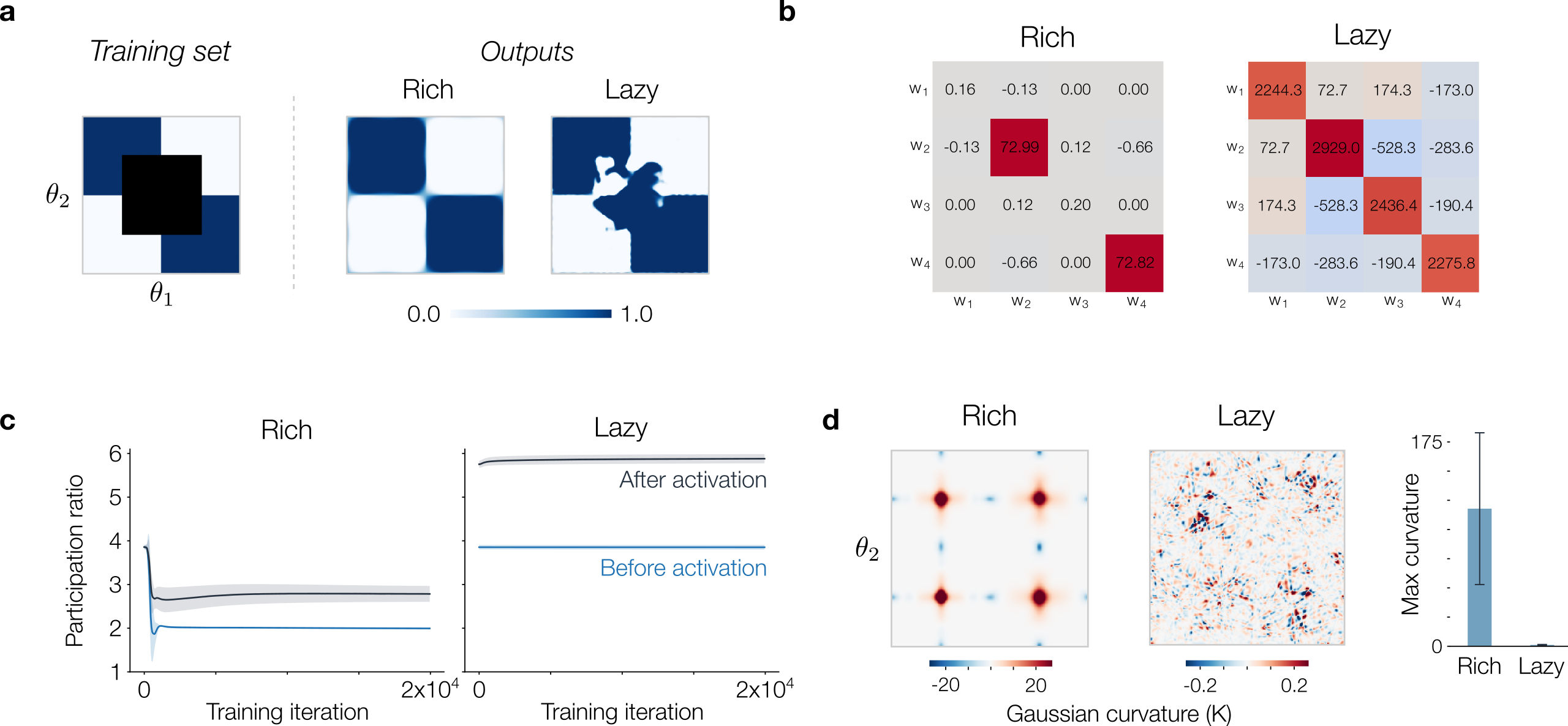}
    \caption{\textbf{Feature learning yields a structured geometry promoting generalisation.} \textbf{a.} Output of rich and lazy networks trained on XOR with a portion of the input manifold held out during training, represented by the black square. Only rich networks are able to generalise to unseen inputs. 
    \textbf{b.} The Gram matrices $W^{\top}W$ of the input weights. Rich networks ignore irrelevant inputs (corresponding to columns). Lazy networks randomly project each input into the high dimensional hidden space. 
    \textbf{c.} Participation ratios over training, with error bars calculated as the standard deviation across 10 different seeds of weight initialisations. Rich networks learn low-dimensional representations, while lazy networks learn high-dimensional representations determined by the initialisation. 
    \textbf{d.} Curvature over the hidden layer manifold, with a bar plot showing the average maximum curvature over the manifold (error bars show standard deviation). Curvature in rich networks is highly structured with large magnitude at class centres. Lazy networks are mostly flat, and have random curvature patterns.}
    \label{fig:richlazy}
    \vspace{-5ex}
\end{figure}

As in previous work, we continuously vary the strength of the initial weights of the network $W_{ij}\sim \mathcal{N}(0, \sigma^2)$. Small initial weights, corresponding to the rich learning regime, tend to generalise better. To test this in our setting, we held out for testing the points on the input manifold $(\theta_1,\theta_2)\in[-\frac{\pi}{2}, \frac{3\pi}{2}] \times [-\frac{\pi}{2}, \frac{3\pi}{2}]$. We found that the rich, but not the lazy learning regime generalised to the unseen points on the manifold (Fig. \ref{fig:richlazy}a). The high-dimensional ($n=100$) rich network learned the optimal low-dimensional projection of the input found by the small-width ($n=4$) network of the previous section, as reflected by its Gram matrix $W^TW$ having sparse low-rank structure (Fig. \ref{fig:richlazy}b, \textit{left}). Instead the lazy network learned random, near-orthogonal projections of the inputs (Fig. \ref{fig:richlazy}b, \textit{right}). 

To better understand the geometry underpinning this generalisation ability, we studied the embedding dimensionality and intrinsic curvature of the representation. First, consistent with previous work, rich learning led to lower-dimensional representations as reflected by the participation ratio $\left(\sum_{i=1}^n\sigma_i\right)^2/\left(\sum_{i=1}^n\sigma_i^2\right)$ of the hidden layer representation of the torus (Fig. \ref{fig:richlazy}c). The discrepancy in the representation was also reflected in the intrinsic geometry. The rich network representation had peaks in positive curvature near the class centres (Fig. \ref{fig:richlazy}d, \textit{left}). These points are where small changes in either input angle lead to a change in hidden layer representation in the same direction, towards switching the value of the discretised input variable. Since the peaks are symmetrical, only being trained on the points on the manifold on one of their halves is sufficient to generalise to the computation to the unseen inputs. In comparison, the lazy network had a mostly flat curvature, with small, randomly spread out, peaks in positive an negative curvature, which did not seem to encode a general computation (Fig. \ref{fig:richlazy}d, \textit{right}). Furthermore, the rich network was more robust to noise in the state-space, likely linked to its highly structured representational geometry, while noise on the manifold had similar effects in both networks  (Fig. \ref{sup_fig2}).

Thus, our results suggest that different learning regimes can have fundamentally different effects on the representational geometry of discrete computations on continuous manifolds. Furthermore, these geometries are tied to different generalisation abilities of the network. 

\begin{figure}[!b]
    \centering
    \includegraphics[width=0.95\linewidth]{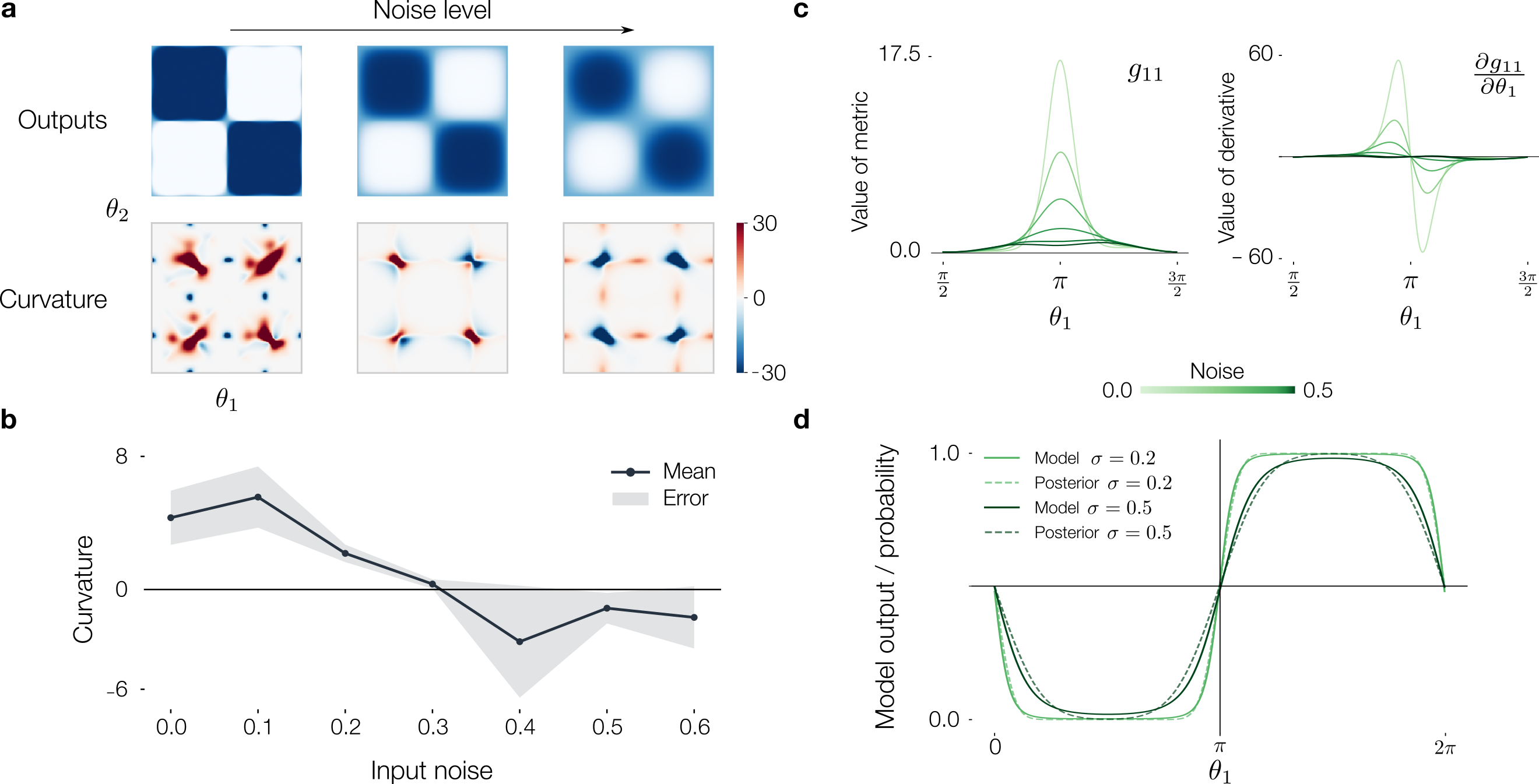}
    \caption{\textbf{The metric and curvature encode the output posterior distribution.} \textbf{a.} Outputs and curvatures of rich networks trained on XOR with different amounts of input noise during training. Increasing noise increases uncertainty near the boundaries, leading to larger regions of outputs close to 0.5. Curvature decreases with noise. \textbf{b.} Mean curvature vs. input noise. Curvature becomes more negative for larger input noises. \textbf{c.} Metric and change in metric in the $\theta_1$ direction for fixed $\theta_2=\frac{\pi}{4}$ for different noises. In noisy models, the metric changes less quickly across the boundaries. \textbf{d.} Outputs learned by models trained with different levels of noise for fixed $\theta_2$ and analytic predictions across the boundary. The learned distribution closely matches the expected posterior distribution and the slope of the posterior is less steep for larger values of noise.}
    \label{fig:noise}
    \vspace{-3ex}
\end{figure}

\subsection{Noise-robust neural representations have low curvature}

Next we asked how injecting noise tangent to the input manifold during training affects the learned representational geometry. We implemented this by a perturbation of the angle $\mathbf{x}=[\cos(\theta_1+\eta_1), \sin(\theta_1+\eta_1), \cos(\theta_2+\eta_2), \sin(\theta_2+\eta_2)]^T$ where $\eta_1,\eta_2$ followed a wrapped normal distribution of variance $\sigma^2$. 

The network output on the original un-noised inputs smoothed out near the class boundaries for networks trained with larger noise levels (Fig. \ref{fig:noise}a, \textit{top}). Interestingly, the curvature near the class centres decreased, even below zero past a certain noise level (Fig. \ref{fig:noise}a, \textit{bottom}; Fig. \ref{fig:noise}b). This suggests that after a certain noise level, small changes in either input angles near the class centres do not lead to similar changes in the representation. Moreover, the value of the metric decreased, near the class boundaries, and so did its change over the manifold, suggesting less warping (Fig. \ref{fig:noise}c). This effect was correlated with the model output learning the flatter posterior distribution of the output (Fig. \ref{fig:noise}d). 

Thus, noise during training affects representational geometry of discrete computations on continuous manifolds. In particular, the continuous decrease in warping with noise reflects the reduced confidence of the network in the output, as quantified by a flatter posterior distribution. Furthermore, this is accompanied by a qualitative change from positive to negative curvature past a certain noise threshold.

% ===== Conclusion =====

\section{Conclusion}

The manifold hypothesis posits that data lie on low-dimensional manifolds. Yet neural networks must often perform fundamentally discrete computations on these continuous manifolds, whether classifying samples from a manifold of images in vision tasks or making discrete actions over continuous state spaces in RL tasks. Here, we sought to understand the effect of this dichotomy on neural network representational geometry. We first showed analytically, in linear networks, that specific Riemannian geometries emerge when learning such discrete computations on continuous manifolds. Then, motivated by this insight, we showed in numerical experiments that the hidden-layer geometry of nonlinear networks reflected a sequential partitioning of these computations into (i) discretisation of continuous variables and (ii) logical computations on the discretised variables. Finally, we demonstrated a link between these effects and different learning regimes, as well as Bayesian computations.

While we focused our analyses on simple networks trained on hand-crafted input manifolds, future work could extend these analyses to more complex architectures trained on real-world data. Indeed, recent work has demonstrated that Riemannian geometry can be used to understand generative \citep{park2023understanding} and vision models \citep{kaul2019riemannian}. In parallel, our toy models could be used to understand further aspects of learning dynamics, such as how the manifold structure of data influences sample complexity \citep{joshi2025learning} or how the covariance between data variables affects generalisation error \citep{loureiro2021learning}.

Furthermore, while our work was mostly descriptive, future work could take a prescriptive approach by manipulating representational geometry via constraints on the metric during training. This could help design models that are (in/equi)variant to specific geometric transformations \citep{tai2019equivariant}, or help place implicit geometric biases into neural networks \citep{katsman2023riemannian}. Finally, some of these analyses could be combined with tools used to infer changes in connectivity \citep{pellegrino2023low} or learning dynamics \citep{confavreux2020meta} from data, in order to study changes in representational geometry over learning in biological networks.

Overall, we show that specific Riemannian geometric structures emerge over learning in networks trained to perform discrete tasks on input data lying on continuous manifolds. These results thus deepen our theoretical understanding of computations in neural networks and their evolution over learning.

% ===== Not counted for page limit =====

\clearpage

\acks{We would like to thank N Alex Cayco-Gajic for providing feedback on an early version of the manuscript.} 

\bibliography{bibliography}

\clearpage

\input{appendix}

\end{document}

%% file: preamble.tex
\usepackage{longtable}% for long tables
\usepackage{lipsum}
\DeclareMathOperator\erf{erf} % for posterior derivation in appendix
\usepackage{booktabs}
\usepackage[load-configurations=version-1]{siunitx} % newer version

\usepackage{soul}
\usepackage{enumitem}
\usepackage{mathtools}

\theorembodyfont{\upshape}
\theoremheaderfont{\scshape}
\theorempostheader{:}
\theoremsep{\newline}

\jmlrvolume{}
\firstpageno{1}
\jmlryear{2025}
\jmlrworkshop{Symmetry and Geometry in Neural Representations}

\title[Discrete computations on continuous manifolds]{Emergent Riemannian geometry over learning\\discrete computations on continuous manifolds}

\newcommand{\supervisorNote}{\thanks{These authors jointly supervised this work}}

   \author{\Name{Julian Brandon} \Email{j.x.brandon@sms.ed.ac.uk}\\%\and
   \addr School of Informatics, University of Edinburgh\\Département d'Etudes Cognitives, Ecole Normale Supérieure PSL\\
   \Name{Angus Chadwick}\supervisorNote \Email{angus.chadwick@ed.ac.uk}\\
   \addr School of Informatics, University of Edinburgh\\
    \Name{Arthur Pellegrino}$^*$ \Email{arthur.pellegrino@ens.fr}\\
  \addr Gastsby Unit, University College London\\Data Science Center, Ecole Normale Supérieure PSL}

%% file: appendix.tex
\appendix

\renewcommand{\thefigure}{S\arabic{figure}}
\renewcommand{\theHfigure}{S\arabic{figure}} 
\setcounter{figure}{0}

\section{Optimal weights for shifted boundaries}\label{apd:weight_rotation}
The toy models considered in this work require networks to learn representations of the input embedding:
\begin{equation}
    x(\theta) = \begin{bmatrix} \cos\theta_1 & \sin\theta_1 & \cos\theta_2 & \sin\theta_2 \end{bmatrix}^\top
\end{equation}
to output a logical operation:
\begin{equation}
    y=\text{LogicOp}(\alpha<\theta_1<\pi+\alpha,\,\alpha<\theta_2<\pi+\alpha)
\end{equation}
where $\alpha$ is a constant used to shift the boundaries.
This is equivalent to solving the task:
\begin{equation}
    y=\text{LogicOP}(v_1>0, v_2>0)
\end{equation}
where $v_1=\sin(\theta_1-\alpha)$ and $v_2=\sin(\theta_2-\alpha)$.
Expanding using the sine addition identity yields 
\begin{equation}
    v_i=\sin\theta_i\cos\alpha - \cos\theta_i\sin\alpha.
\end{equation}
Note that $\mathbf{x}^{(2i-1)}=\cos\theta_i$ and $\mathbf{x}^{(2i)}=\sin\theta_i$. As was shown in Section~\ref{sec:learning_dynamics}, to efficiently solve the task the hidden layer must learn a mapping:
\begin{align}
    h(\mathbf{x}) &= v_1+v_2 \\
    &= \underbrace{-\sin \alpha}_{\text{weight for } \mathbf{x}^{(1)}} \cdot \mathbf{x}^{(1)} + 
\underbrace{\cos \alpha}_{\text{weight for } \mathbf{x}^{(2)}} \cdot \mathbf{x}^{(2)} + 
\underbrace{-\sin \alpha}_{\text{weight for } \mathbf{x}^{(3)}} \cdot \mathbf{x}^{(3)} + 
\underbrace{\cos \alpha}_{\text{weight for } \mathbf{x}^{(4)}} \cdot \mathbf{x}^{(4)}\label{eq:rotated_weights}
\end{align}
Collecting the terms from Eq.~\ref{eq:rotated_weights} reveals the structure of the optimal weight vector $\mathbf{w}$. For each input pair $(\cos\theta_i,\sin\theta_i)$, the corresponding weights are $(-\sin\alpha,\cos\alpha)$. We observe that this weight vector is the result of applying a rotation matrix $R(\alpha)$ to the unshifted weights (where $\alpha=0$):

\begin{equation}
\begin{pmatrix}
-\sin\alpha \\
\cos\alpha
\end{pmatrix} = \begin{pmatrix}
\cos\alpha & -\sin\alpha\\
\sin\alpha & \cos\alpha
\end{pmatrix}  \begin{pmatrix}
    0 \\
    1
\end{pmatrix}
\end{equation}

Consequently, shifting the classification boundary by $\alpha$ in the input space is mathematically equivalent to rotating the optimal hidden layer weights by $\alpha$ in the parameter space.

\section{Linear network learning dynamics}\label{apd:saxe_dynamics}
\subsection*{Input Embedding and Target Function (AND Task)}
The angles are embedded into a 4-dimensional input vector $x \in \mathbb{R}^4$:
\begin{equation}
x(\theta) = \begin{bmatrix} \cos\theta_1 & \sin\theta_1 & \cos\theta_2 & \sin\theta_2 \end{bmatrix}^T
\end{equation}
The target $y$ implements a logical AND on the half-planes defined by the angles. We define the active region as the first quadrant $[0, \pi] \times [0, \pi]$:
\begin{equation}
y(\theta) = 
\begin{cases} 
1 & \text{if } 0 < \theta_1 < \pi \text{ AND } 0 < \theta_2 < \pi \\
0 & \text{otherwise}
\end{cases}
\end{equation}

\subsection*{Statistical Moments}

The learning dynamics of a deep linear network are governed by the input correlation matrix $\Sigma^{11}$ and the input-output correlation matrix $\Sigma^{31}$. Since we minimize MSE, these are defined as expectations over the manifold.

\subsection*{Input Correlation Matrix $\Sigma^{11}$}
We calculate $\Sigma^{11} = \mathbb{E}[\mathbf{x}\mathbf{x}^T]$. Due to the orthogonality of sine and cosine functions, the off-diagonal elements vanish. Assuming a uniform sampling distribution $p(\theta)=\frac{1}{2\pi}$, we compute the diagonal variance for a sine term:
\begin{equation}
\sigma^2 = \mathbb{E}[\sin^2\theta] = \int_{-\pi}^{\pi} \int_{-\pi}^{\pi} \sin^2(\theta_1) p(\theta) \, d\theta_1 d\theta_2
\end{equation}
\begin{equation}
\sigma^2 = \frac{1}{2\pi} \int_{-\pi}^{\pi} \sin^2(\theta_1) \, d\theta_1 = \frac{1}{2\pi} (\pi) = 0.5
\end{equation}
Thus, the input correlation matrix is diagonal:
\begin{equation}
\Sigma^{11} = 0.5 I
\end{equation}
In the notation of Saxe et al. (Appendix E), the eigenvalue matrix is $D = 0.5 I$.

\subsection*{Input-Output Correlation Matrix $\Sigma^{31}$}
We calculate $\Sigma^{31} = \mathbb{E}[yx^T]$. The integration is restricted to the active region where $y=1$.
\begin{equation}
\Sigma^{31} = \int_{0}^{\pi} \int_{0}^{\pi} x(\theta)^T p(\theta) \, d\theta_1 d\theta_2
\end{equation}
Evaluating the component for $\sin\theta_1$:
\begin{equation}
\mathbb{E}[y \sin\theta_1] = \frac{1}{4\pi^2} \left( \int_{0}^{\pi} \sin\theta_1 d\theta_1 \right) \left( \int_{0}^{\pi} 1 d\theta_2 \right)
\end{equation}
\begin{equation}
= \frac{1}{4\pi^2} (2)(\pi) = \frac{2\pi}{4\pi^2} = \frac{1}{2\pi}
\end{equation}
By symmetry, the expectation for $\sin\theta_2$ is identical. The cosine terms integrate to zero over $[0, \pi]$. The correlation vector is:
\begin{equation}
\Sigma^{31} = \begin{bmatrix} 0 & \frac{1}{2\pi} & 0 & \frac{1}{2\pi} \end{bmatrix}
\end{equation}
The non-zero singular value $s$ corresponds to the Euclidean norm of this vector:
\begin{equation}
s = \sqrt{\left(\frac{1}{2\pi}\right)^2 + \left(\frac{1}{2\pi}\right)^2} = \frac{\sqrt{2}}{2\pi} = \frac{1}{\sqrt{2}\pi}
\end{equation}
which has corresponding right singular vector
\begin{equation}
    \mathbf{v}_1=\begin{bmatrix}0 & \frac{1}{\sqrt{2}} & 0 & \frac{1}{\sqrt{2}}\end{bmatrix}^\top
\end{equation}

\subsection*{Differential Equation for Weight Magnitude}
Following Saxe et al. (2013), the gradient descent dynamics for the weights $W$ are given by:
\begin{equation}
\tau \frac{d}{dt} W = \Sigma^{31} - W \Sigma^{11}
\end{equation}
We project these dynamics onto the single active mode. Let 
\begin{equation}\label{eq:u_t_def}
    u(t)=W_2(t)W_1(t)\mathbf{v}_1
\end{equation} 
be the scalar product of the input and output weights aligned with the mode. Assuming balanced initialization ($W_1(0)\mathbf{v}_1 \approx W_2(0)^\top$), the chain rule introduces a factor of 2:
\begin{equation}
\tau \frac{du}{dt} = 2u (s - u \cdot \sigma^2)
\end{equation}
Substituting $\sigma^2 = 0.5$:
\begin{equation}
\tau \frac{du}{dt} = 2u (s - 0.5 u)
\end{equation}
Rearranging to group the constants:
\begin{equation}
\tau \frac{du}{dt} = u (2s - u)
\end{equation}
Let $S = 2s$ be the effective singular value. This constant represents the final asymptotic strength of the weights.
\begin{equation}
\label{eq:logistic}
\tau \frac{du}{dt} = u (S - u)
\end{equation}

\subsection*{Integration and Solution}

We solve the differential equation \eqref{eq:logistic} by separation of variables:
\begin{equation}
\int \frac{du}{u(S - u)} = \int \frac{dt}{\tau}
\end{equation}
Using partial fraction decomposition $\frac{1}{u(S-u)} = \frac{1}{S}(\frac{1}{u} + \frac{1}{S-u})$:
\begin{equation}
\frac{1}{S} \left( \ln u - \ln(S-u) \right) = \frac{t}{\tau} + C'
\end{equation}
\begin{equation}
\ln \left( \frac{u}{S-u} \right) = \frac{S t}{\tau} + C
\end{equation}
Exponentiating and solving for $u(t)$ yields the sigmoidal learning curve:
\begin{equation}\label{eq:u_t_dynamics}
u(t) = \frac{S e^{St/\tau}}{e^{St/\tau} - 1 + \frac{S}{u_0}}
\end{equation}

\subsection*{Learned Weights}
To obtain the form of the learned hidden layer weights, we first project the hidden weights onto the learned mode:
\begin{equation}\label{eq:a_def}
    \mathbf{a}= \begin{bmatrix}a_1 & a_2 & a_3 & a_4\end{bmatrix}^\top \coloneqq W_1\mathbf{v}_1
\end{equation}
To recover the learned weights we perform the opposite transformation:
\begin{equation}
    W_1^{\text{learned}}=\mathbf{a}\mathbf{v}_1^\top=\frac{1}{\sqrt{2}}\begin{bmatrix}
    0 & a_1 & 0 & a_1 \\
    0 & a_2& 0 & a_2 \\ 
    0 & a_3 & 0 & a_3 \\
    0 & a_4 & 0 & a_4
    \end{bmatrix}
\end{equation}

\subsection*{Pullback Metric Dynamics}
To derive the dynamics of the hidden pullback metric over training, we first define the Jacobian of the hidden activations with respect to task angles:
\begin{equation}
    J_\mathbf{z}=\frac{\partial\mathbf{z}}{\partial\theta} 
    =\frac{\partial\mathbf{z}}{\partial\mathbf{x}}\frac{\partial\mathbf{x}}{\partial\theta} 
    = W_1J_\mathbf{x}
\end{equation}
The hidden pullback metric is then:
\begin{align}\label{eq:hidden_metric}
    G_\mathbf{z}&=J_\mathbf{z}^\top J_\mathbf{z} \\
    &= J_\mathbf{x}^\top W_1^\top W_1 J_\mathbf{x}
\end{align}
For randomly initialised hidden weights, we can decompose into weights parallel and perpendicular to the learned mode: $W_1(t)=W_\parallel(t) + W_\perp=\mathbf{a}(t)\mathbf{v}_1^\top+W_\perp$, where $\mathbf{a}(t)$ and $\mathbf{v}_1$ are defined as above. The perpendicular weights have zero gradients, and so don't evolve over training.
The metric then becomes:
\begin{align}
    G_\mathbf{z}&=J_\mathbf{x}^\top (W_\parallel+W_\perp)^\top (W_\parallel + W_\perp)J_\mathbf{x}\label{eq:metric_decomp} \\
    &= J_\mathbf{x}^\top (\mathbf{a}(t)\mathbf{v}_1^\top)^\top (\mathbf{a}(t)\mathbf{v}_1^\top)J\mathbf{x} + G_\mathbf{z}^\perp \\
    &=\frac{1}{2}a(t)^2J_\mathbf{x}^\top \mathbf{v}_1\mathbf{v}_1^\top J_\mathbf{x} + G_\mathbf{z}^\perp \label{eq:metric_a} \\
    &= \frac{1}{2}a(t)^2 \begin{pmatrix}
    \cos^2\theta_1 & \cos\theta_1\cos\theta_2 \\
    \cos\theta_1\cos\theta_2 & \cos^2\theta_2
    \end{pmatrix} +G_\mathbf{z}^\perp \\
    &=\frac{1}{2}a(t)^2G_\mathbf{z}^\text{task} + G_\mathbf{z}^\perp
\end{align}
where in (\ref{eq:metric_a}) we have used the fact that $\mathbf{a}(t)=a(t)\cdot \hat{\mathbf{u}}$, and $\hat{\mathbf{u}}$ is a constant unit vector. In the rich regime ($\sigma^2\ll 1$) $G_\mathbf{z}^\perp$ is small. Note that cross terms in the expansion of (\ref{eq:metric_decomp}) vanish due to orthogonality. Over learning, $a(t)^2$ grows and the final learned metric is approximately:
\begin{equation}
    G_\mathbf{z}\simeq \frac{1}{2}a(t)^2 \begin{pmatrix}
    \cos^2\theta_1 & \cos\theta_1\cos\theta_2 \\
    \cos\theta_1\cos\theta_2 & \cos^2\theta_2
    \end{pmatrix}
\end{equation}
% From previous analysis, $W_1$ evolves according to the learned mode $a_1(t)$ so we can write the only learned hidden mode as $W_1^{(1)}=a_1(t)\mathbf{v}_1^\top$. Substituting this expression into Eq.~\ref{eq:hidden_metric}:
% \begin{align}
%     G_\mathbf{z}&=J_\mathbf{x}\left(a_1(t)\mathbf{v}_1^\top\right)^\top\left( a_1(t)\mathbf{v}^\top\right)J_\mathbf{x} \\
%     &=a_1(t)^2 J_\mathbf{x}^\top \mathbf{v}_1\mathbf{v}_1^\top J_\mathbf{x} \\
%     &\simeq \frac{1}{2}u(t)\begin{pmatrix}
%     \cos^2\theta_1 & \cos\theta_1\cos\theta_2 \\
%     \cos\theta_1\cos\theta_2 & \cos^2\theta_2
%     \end{pmatrix}\label{eq:final_G_dynamics}
% \end{align}
% where in (\ref{eq:final_G_dynamics}) we have used the fact that $a(t)\simeq\sqrt{u(t)}$ and substituted the form of $J_\mathbf{x}$ for our torus embedding of inputs.

\begin{figure}
    \centering
    \includegraphics[width=0.4\linewidth]{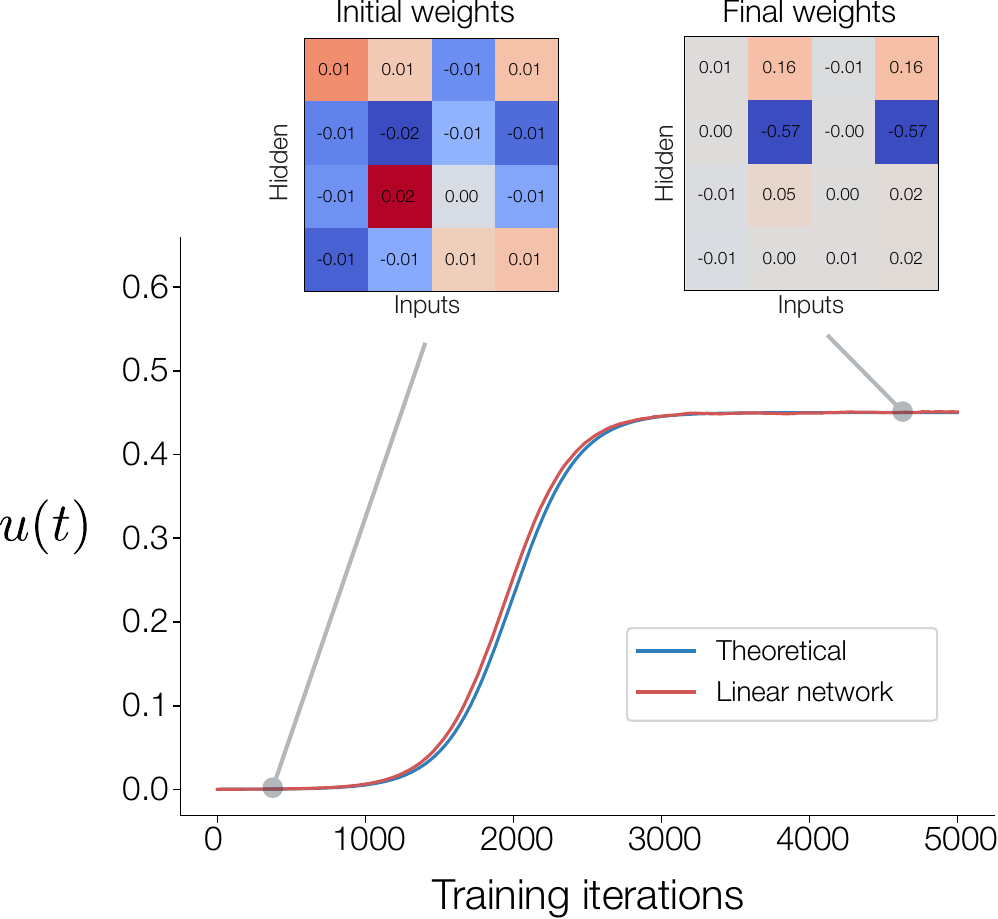}
    \caption{\textit{Left}: Theoretical curve of learning dynamics $u(t)$ plotted against a linear and tanh network trained on the AND task. The theoretical curve is an excellent prediction of learning dynamics in the linear network, and is a reasonable approximation of the initial learning stage in the non-linear network. \textit{Right}: Learned hidden layer weights for the tanh network. Only weights corresponding to the non-zero mode are learned.}
    \label{fig:saxe_learning_dynamics}
\end{figure}

\section{Geometry of Computations in Logic Gate Tasks}\label{apd:first}

\subsection{1 Layer XOR vs 2 Layer XOR}
\begin{figure}[h!]\label{fig:app_1vs2_XOR}
    \centering
    \includegraphics[width=1\linewidth]{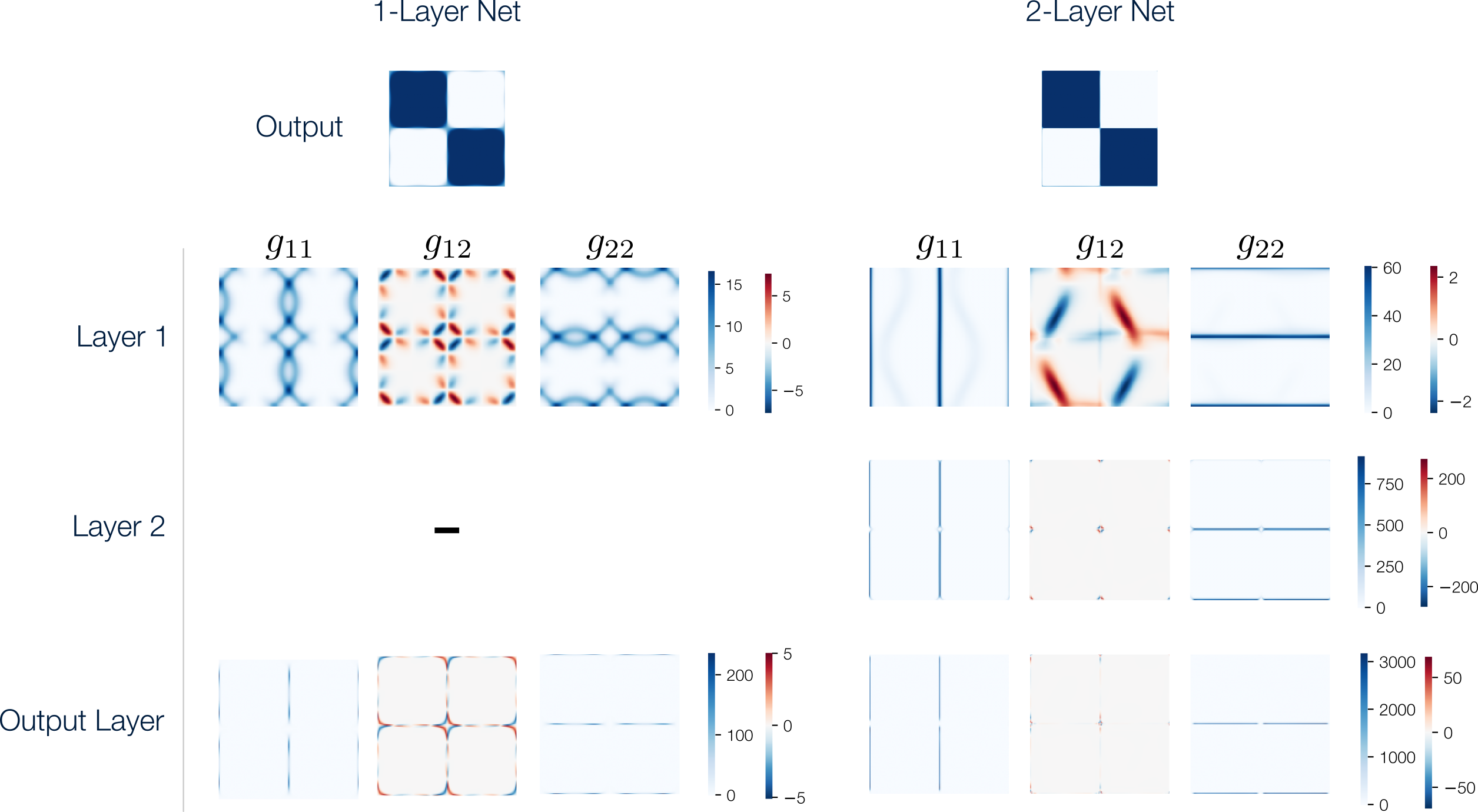}
    \caption{\textbf{Varying depth of XOR networks} \textit{Left}: Hidden and output metrics for the 1 hidden layer XOR network. \textit{Right}: Hidden and output metrics for the 2 hidden layer XOR network.}
    \label{sup_fig1}
\end{figure}

We trained networks on the XOR task with 1 and 2 layers and a flat torus input manifold. The 2 layer network (Fig.~\ref{fig:app_1vs2_XOR}, \textit{right}) decomposed input-manifold specific computations and task-specific computations across layers, with the first hidden layer metric showing the characteristic discretisation pattern found in AND and OR networks (Fig.~\ref{fig:fig1}e). The 1-layer hidden metric shows additional structure with sensitivity oscillating across the discretisation boundaries, suggesting a more complex computation is occurring. The XOR task with discrete inputs requires at least one hidden layer to solve \citep{minsky1969introduction}, so the 1-layer network must discretise and solve the logic gate in one ``step''.

\subsection{Analytic Expression for Hidden Metric in XOR Task}
Consider a network trained on the XOR task with a tanh hidden non-linearity and input weights $W\in\mathbb{R}^{K\times 4}$ with components $w_{a,b}$ for the weight between the $b^{\text{th}}$ input and the $a^{\text{th}}$ hidden neuron.
Inputs $\theta_1$ and $\theta_2$ are embedded on a flat torus
\begin{equation}\label{eq:input_embedding}
    \mathbf{x}=[\cos(\theta_1),\sin(\theta_1),\cos(\theta_2),\sin(\theta_2)]^\top,
\end{equation} 
such that the target outputs are $\text{XOR}(0\leq\theta_1<\pi,0\leq\theta_2<\pi)$. The hidden layer activations are $\mathbf{z}=\tanh(\mathbf{u})=\tanh(W\mathbf{x})\in\mathbb{R}^{K}$. We first calculate the Jacobian of the hidden activations with respect to the inputs:
\[
\begin{split}
    J_\mathbf{z}&=\left[ \frac{\partial\mathbf{z}}{\partial\theta_1}\,\,\,\frac{\partial\mathbf{z}}{\partial\theta_2}\right] \\
    & = \left[ \frac{d\mathbf{z}}{d\mathbf{u}}\frac{\partial\mathbf{u}}{\partial\theta_1}\,\,\, \frac{d\mathbf{z}}{d\mathbf{u}}\frac{\partial\mathbf{u}}{\partial\theta_2}\right] \\ 
    & = \text{diag}(\text{sech}^2(\mathbf{u}))\left[ \frac{d\mathbf{u}}{d\mathbf{x}}\frac{\partial\mathbf{x}}{\partial\theta_1}\,\,\, \frac{d\mathbf{u}}{d\mathbf{x}}\frac{\partial\mathbf{x}}{\partial\theta_2}\right] \\
        &= \text{diag}(\text{sech}^2(W\mathbf{x}))WJ_{\mathbf{x}}
        \end{split}
\]
where from Equation~\ref{eq:input_embedding} we calculate:
\begin{equation}
    J_{\mathbf{x}}=
    \begin{bmatrix}
    -\sin(\theta_1) & 0 \\
    \cos(\theta_1) & 0 \\
    0 & -\sin(\theta_2) \\
    0 & \cos(\theta_2)
    \end{bmatrix}.
\end{equation}
The metric tensor of the hidden layer is thus:
\begin{equation}
\begin{split}
    G_{\mathbf{z}}&=J_\mathbf{z}^\top J_\mathbf{z} \\
    &= J_\mathbf{x}^\top W^\top \text{diag}(\text{sech}^4(W\mathbf{x}))WJ_\mathbf{x}.
    \end{split}
\end{equation}
Which can be written in component form as:
\begin{equation}\label{eq:G_components}
\begin{split}
    G_{ij}&=\left(J_\mathbf{x}^\top W^\top DWJ_\mathbf{x}\right)_{ij} \\
    &=\sum_{k=1}^K(WJ_\mathbf{x})_{ki}(WJ_\mathbf{x})_{kj}D_{kk}
    \end{split}
\end{equation} where $D=\text{diag}(\text{sech}^4(W\mathbf{x}))$.
Assume $w_{k1}=w_{k3}=0$, then
\begin{equation}\label{eq:WJ_def}
    (WJ_\mathbf{x})_{k1}=w_{k2}\cos(\theta_1), \,\,\,(WJ_\mathbf{x})_{k2}=w_{k4}\cos(\theta_2)
\end{equation}
and the pre-activations are
\begin{equation}\label{eq:u_def}
u_k=(W\mathbf{x})_k=w_{k2}\sin(\theta_1)+w_{k4}\sin(\theta_2).
\end{equation}
Substitute (\ref{eq:WJ_def})-(\ref{eq:u_def}) into (\ref{eq:G_components}):
\begin{equation}\label{eq:post_act_metric}
\begin{split}
    G_{ij} &= \sum_{k=1}^K (WJ_\mathbf{x})_{ki}(WJ_\mathbf{x})_{kj}\,\text{sech}^4(u_k) \\
    &= \sum_{k=1}^K (w_{k,2i}\cos(\theta_i))(w_{k,2j}\cos(\theta_j))\text{sech}^4[w_{k2}\sin(\theta_1)+w_{k4}\sin(\theta_2)] \\
    &= \cos(\theta_i)\cos(\theta_j)\sum_{k=1}^Kw_{k,2i}w_{k,2j}\text{sech}^4[w_{k2}\sin(\theta_1)+w_{k4}\sin(\theta_2)] \\
    &= \cos(\theta_i)\cos(\theta_j)g^{(ij)}(\theta_1,\theta_2)
\end{split}
\end{equation}
where $2i\in\{2,4\}$ corresponds to the input index of $\theta_i$ ($i=1\rightarrow2,i=2\rightarrow4)$. 

To observe how the Gaussian curvature depends on this expression, we use the Brioschi formula which has the form:
\begin{equation}\label{eq:brioschi_formula}
    K=\frac{M(\theta_1,\theta_2)}{(\text{det(G)})^2}.
\end{equation}
Substituting (\ref{eq:post_act_metric}) into (\ref{eq:brioschi_formula}) yields:
\begin{equation}
\begin{split}
    K &= \frac{M(\theta_1, \theta_2)}{\left[\cos^2(\theta_1)\cos^2(\theta_2)g^{(11)}(\theta_1,\theta_2)g^{(22)}(\theta_1,\theta_2)-\cos^2(\theta_1)\cos^2(\theta_2)g^{(12)}(\theta_1,\theta_2)^2\right]^2} \\
    &= \frac{1}{\cos^4(\theta_1)\cos^4(\theta_2)} 
    \frac{M(\theta_1,\theta_2)}{\left[g^{(11)} g^{(22)} - (g^{(12)})^2\right]^2} \\
    &= \frac{M'(\theta_1, \theta_2)}{\cos^4(\theta_1)\cos^4(\theta_2)}
    \end{split}
\end{equation}

\section{Rich and lazy learning geometry}

\begin{figure}
    \centering
    \includegraphics[width=0.9\linewidth]{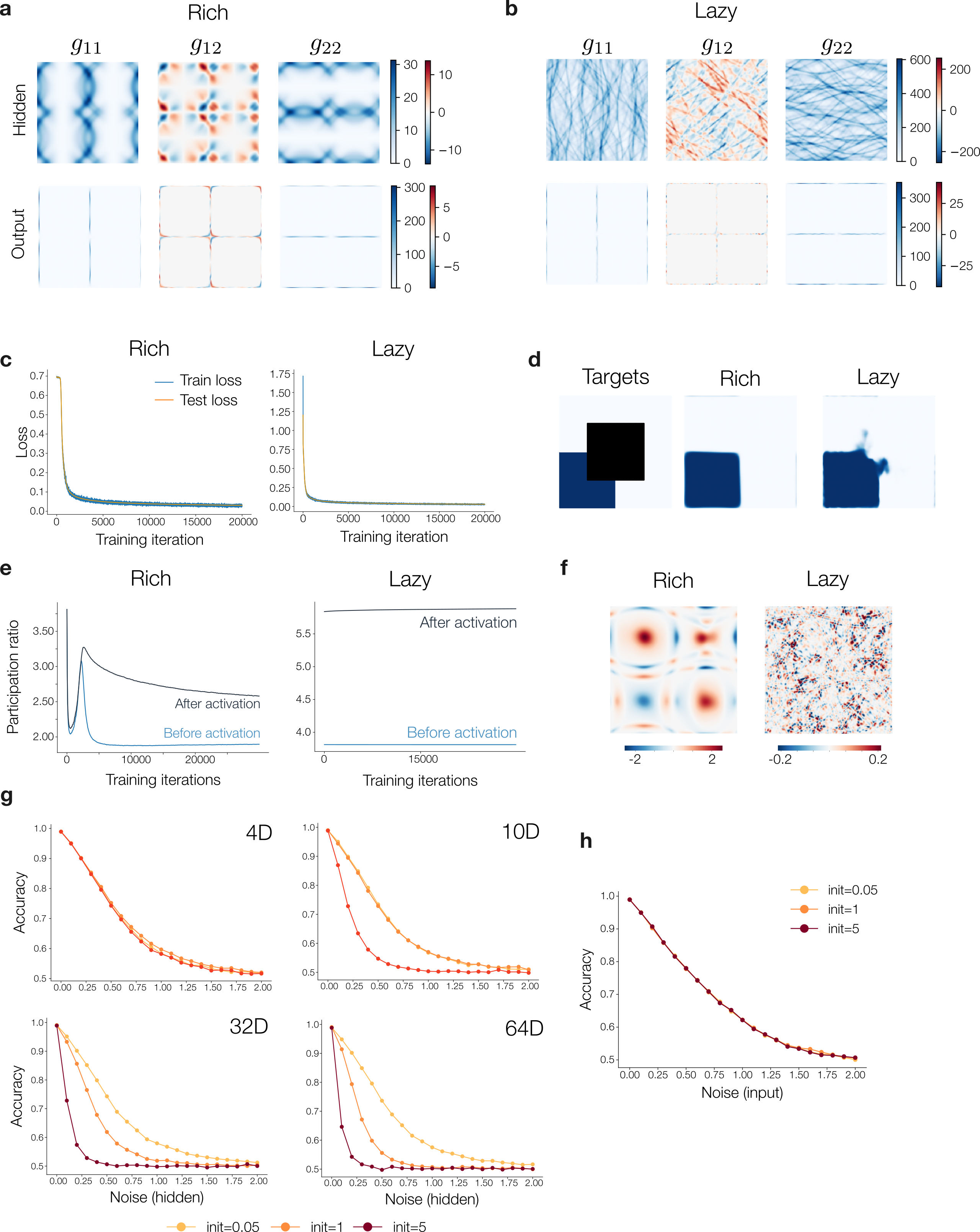}
    \caption{\textbf{Rich and lazy geometry} \textbf{a.} The hidden and output metric components for the XOR network in the rich regime. \textbf{b.} Metric components for the lazy XOR network. \textbf{c.} Loss curves for rich and lazy XOR networks. \textbf{d.} Outputs of rich and lazy AND networks trained with a portion of the input space held out during training. Rich networks generalise better to unseen inputs than lazy networks. \textbf{e.} Participation ratios of rich and lazy AND networks across training. \textbf{f.} The hidden layer Gaussian curvature across the manifold in rich and lazy networks. \textbf{g.} Accuracy as a function of variance of input embedding noise for networks trained with varying weight initialisations and increasing input embedding dimension. \textbf{h.} Accuracy as a function of variance of task variable noise for varying weight initialisations.}
    \label{sup_fig2}
\end{figure}

Rich networks trained on XOR had low-dimensional hidden manifolds, and structured Gram matrices, whereas lazy networks had high-dimensional hidden manifolds and near-orthogonal projections of inputs. This difference was also captured by the hidden layer metrics. Rich networks had localised sensitivity to inputs near the boundaries (Fig.~\ref{sup_fig2}a), converging to the small width solutions (Fig.~\ref{sup_fig1}, \textit{left}), whereas lazy networks had random metrics with large magnitude across the manifold (Fig~\ref{sup_fig2}b). Output metrics in both networks showed sensitivity close to task boundaries, indicating that a good solution was found. The loss curves of the rich network show a slight initial delay in learning compared to the lazy network (Fig.\ref{sup_fig2}) consistent with the prediction of a slower convergence in rich networks \citep{saxe2013exact}, although this effect is small.

We also analysed the geometry of rich and lazy networks in the AND task, observing better generalisation (Fig~\ref{sup_fig2}d), a lower-dimensional hidden manifold (Fig~\ref{sup_fig2}e), and more structured and larger curvature (Fig~\ref{sup_fig2}f), consistent with the XOR results. The rich network's curvature was negative in one quadrant and positive in the other three, mirroring the target outputs. This suggests the curvature pulls apart unlike classification regions on the hidden manifold.

We tested the robustness to noise of XOR networks for a range of weight initialisations (Fig~\ref{sup_fig2}g). Noise at the level of the task variables $\theta_1,\theta_2$ showed no difference in robustness between learning regimes (\ref{sup_fig2}h), because the mapping from task variables to outputs was approximately identical irrespective of internal geometry. Isotropic noise at the level of embedded inputs $\mathbf{x}(\theta_1,\theta_2)$ showed differences in robustness (Fig~\ref{sup_fig2}g). Here, task were embedded first on a 4-D flat torus, and then by an orthogonal projection to a higher-dimensional space. The difference in robustness increased with embedding dimension. High-dimensional spaces containing low-dimensional manifolds have a large number of directions orthogonal (task-irrelevant) to the manifold, which increases with embedding dimension. Rich networks are known to compress task-irrelevant dimensions \citep{paccolatGeometricCompressionInvariant2021} and so become increasingly robust to noise off the manifold as the number of orthogonal dimensions increases.

\section{A Bayesian Model of Noise}
To study whether networks learn approximately Bayesian inference under training noise, we computed the posterior for a simpler 1-dimensional classification problem on a circle. 

We consider a random variable $\delta$ uniformly distributed on a circle, representing an angle in $[-\pi, \pi)$. We observe a noisy measurement $c = \delta + \eta$, where the noise $\eta$ is drawn from a zero-mean Gaussian distribution, $\eta \sim \mathcal{N}(0, \sigma^2)$.

Our goal is to determine the probability that $\delta$ is in the upper semi-circle, $[0, \pi)$, given the measurement $c$. We define a binary variable $A$:
\begin{equation*}
    A = 
    \begin{cases} 
        +1 & \text{if } 0 \le \delta < \pi \\
        -1 & \text{if } -\pi \le \delta < 0 
    \end{cases}
\end{equation*}
We want to find the posterior probability $P(A=1|c)$. From Bayes' theorem, and given that the priors $P(A=1)$ and $P(A=-1)$ are both $0.5$, the posterior is:
\begin{equation*}
    P(A=1|c) = \frac{P(c|A=1)}{P(c|A=1) + P(c|A=-1)}
\end{equation*}

\paragraph{Likelihoods}
For a circular variable, the Gaussian noise wraps around the circle. The conditional probability of observing $c$ given $\delta$ is given by the wrapped normal distribution:
\begin{equation*}
    P(c|\delta) = \sum_{k=-\infty}^{\infty} \frac{1}{\sqrt{2\pi\sigma^2}} \exp\left(-\frac{(c - \delta - 2\pi k)^2}{2\sigma^2}\right)
\end{equation*}

\paragraph{Likelihood for A=1}
We find $P(c|A=1)$ by marginalizing over $\delta \in [0, \pi)$. Given $A=1$, the PDF $p(\delta|A=1) = 1/\pi$.
\begin{align*}
    P(c|A=1) &= \int_{0}^{\pi} P(c|\delta) p(\delta|A=1) d\delta \\
    &= \frac{1}{\pi} \int_{0}^{\pi} \sum_{k=-\infty}^{\infty} \frac{1}{\sqrt{2\pi\sigma^2}} \exp\left(-\frac{(c - \delta - 2\pi k)^2}{2\sigma^2}\right) d\delta \\
    % Swap integral and summation
    &= \frac{1}{\pi\sqrt{2\pi\sigma^2}} \sum_{k=-\infty}^{\infty} \int_{0}^{\pi} \exp\left(-\frac{(c - \delta - 2\pi k)^2}{2\sigma^2}\right) d\delta
\end{align*}
The integral of a Gaussian is related to the error function, $\erf(z)$. Evaluating the integral gives:
\begin{align*}
    P(c|A=1) &= \frac{1}{2\pi} \sum_{k=-\infty}^{\infty} \left[ \erf\left(\frac{\delta - c + 2\pi k}{\sqrt{2}\sigma}\right) \right]_0^\pi \\
    &= \frac{1}{2\pi} \sum_{k=-\infty}^{\infty} \left[ \erf\left(\frac{\pi - c + 2\pi k}{\sqrt{2}\sigma}\right) - \erf\left(\frac{-c + 2\pi k}{\sqrt{2}\sigma}\right) \right] \\
    % Using erf(-z) = -erf(z)
    &= \frac{1}{2\pi} \sum_{k=-\infty}^{\infty} \left[ \erf\left(\frac{\pi - c + 2\pi k}{\sqrt{2}\sigma}\right) + \erf\left(\frac{c - 2\pi k}{\sqrt{2}\sigma}\right) \right]
\end{align*}

\paragraph{Likelihood for A=-1}
Similarly, for $A=-1$, we marginalize over $\delta \in [-\pi, 0)$, where $p(\delta|A=-1) = 1/\pi$.
\begin{align*}
    P(c|A=-1) &= \frac{1}{\pi} \int_{-\pi}^{0} \sum_{k=-\infty}^{\infty} \frac{1}{\sqrt{2\pi\sigma^2}} \exp\left(-\frac{(c - \delta - 2\pi k)^2}{2\sigma^2}\right) d\delta \\
    &= \frac{1}{2\pi} \sum_{k=-\infty}^{\infty} \left[ \erf\left(\frac{\delta - c + 2\pi k}{\sqrt{2}\sigma}\right) \right]_{-\pi}^0 \\
    &= \frac{1}{2\pi} \sum_{k=-\infty}^{\infty} \left[ \erf\left(\frac{-c + 2\pi k}{\sqrt{2}\sigma}\right) - \erf\left(\frac{-\pi - c + 2\pi k}{\sqrt{2}\sigma}\right) \right] \\
    &= \frac{1}{2\pi} \sum_{k=-\infty}^{\infty} \left[ \erf\left(\frac{\pi + c - 2\pi k}{\sqrt{2}\sigma}\right) - \erf\left(\frac{c - 2\pi k}{\sqrt{2}\sigma}\right) \right]
\end{align*}

\paragraph{Final Posterior Probability}
Substituting the likelihoods into the posterior formula, the $\frac{1}{2\pi}$ factor cancels. The denominator is the sum of the numerators of the two likelihoods.
\begin{align*}
    \text{Numerator} &= \sum_{k=-\infty}^{\infty} \left[ \erf\left(\frac{\pi - c + 2\pi k}{\sigma\sqrt{2}}\right) + \erf\left(\frac{c - 2\pi k}{\sigma\sqrt{2}}\right) \right] \\
    \text{Denominator} &= \sum_{k=-\infty}^{\infty} \left[ \left( \erf\left(\frac{\pi - c + 2\pi k}{\sigma\sqrt{2}}\right) + \erf\left(\frac{c - 2\pi k}{\sigma\sqrt{2}}\right) \right) \right. \\ 
    & \qquad \left. + \left( \erf\left(\frac{\pi + c - 2\pi k}{\sigma\sqrt{2}}\right) - \erf\left(\frac{c - 2\pi k}{\sigma\sqrt{2}}\right) \right) \right] \\
    &= \sum_{k=-\infty}^{\infty} \left[ \erf\left(\frac{\pi - c + 2\pi k}{\sigma\sqrt{2}}\right) + \erf\left(\frac{\pi + c - 2\pi k}{\sigma\sqrt{2}}\right) \right]
\end{align*}
The final expression for the posterior probability is:
\begin{equation*}
    P(A=1|c) = \frac{\sum_{k=-\infty}^{\infty} \left[ \erf\left(\frac{\pi - c + 2\pi k}{\sigma\sqrt{2}}\right) + \erf\left(\frac{c - 2\pi k}{\sigma\sqrt{2}}\right) \right]}{\sum_{k=-\infty}^{\infty} \left[ \erf\left(\frac{\pi - c + 2\pi k}{\sigma\sqrt{2}}\right) + \erf\left(\frac{\pi + c - 2\pi k}{\sigma\sqrt{2}}\right) \right]}
\end{equation*}
where range of the index $k$ controls how many ``wrap-arounds'' of the circle are considered. The term in the sum corresponding to $k=0$ is equivalent to the posterior without periodic boundaries (i.e. a line). The circular posterior for $k=\{-1,0,1\}$, alongside the distribution learned by the network is plotted in Figure~\ref{fig:noise}d.